\documentclass[acmlarge]{acmart}
\copyrightyear{2026}
\acmYear{2026}
\setcopyright{cc}
\setcctype{by}
\acmConference[FAccT '26]{The 2026 ACM Conference on Fairness, Accountability, and Transparency}{June 25--28, 2026}{Montreal, QC, Canada}
\acmBooktitle{The 2026 ACM Conference on Fairness, Accountability, and Transparency (FAccT '26), June 25--28, 2026, Montreal, QC, Canada}
\acmDOI{10.1145/3805689.3806504}
\acmISBN{979-8-4007-2596-8/2026/06}
\makeatletter
\newcommand{\myconfshort}{\acmConference@shortname}
\newcommand{\myconffull}{\acmConference@name}
\newcommand{\myconfdate}{\acmConference@date}
\newcommand{\myconfloc}{\acmConference@venue}
\AtBeginDocument{
  \fancypagestyle{firstpagestyle}{
    \fancyhead{}%
    \fancyfoot[C]{}%
  }
  \fancyhf{}
  \fancyhead[LO]{\@headfootfont\shorttitle}%
  \fancyhead[RE]{\@headfootfont\@shortauthors}%
  \fancyhead[LE]{\@headfootfont\footnotesize \myconfshort, \myconfdate, \myconfloc}%
  \fancyhead[RO]{\@headfootfont\footnotesize \myconfshort, \myconfdate, \myconfloc}%
  \fancyfoot[C]{}%
}
\makeatother
\AtBeginDocument{\settopmatter{printacmref=true}}
\usepackage{array}
\usepackage{enumitem}
\AtBeginDocument{%
  }

\begin{document}

\title{Standard Language Ideology in AI-Generated Language}

\author{Genevieve Smith*}
\email{genevieve.smith@stanford.edu}
\affiliation{%
  \institution{Stanford University}
  \city{Palo Alto}
  \state{CA}
  \country{USA}
}

\author{Eve Fleisig*}
\email{efleisig@berkeley.edu}
\affiliation{%
  \institution{UC Berkeley}
  \city{Berkeley}
  \state{CA}
  \country{USA}
}

\author{Ishita Rustagi}
\affiliation{%
  \institution{UC Berkeley}
  \city{Berkeley}
  \state{CA}
  \country{USA}
}

\author{Xavier Yin}
\affiliation{%
  \institution{UC Berkeley}
  \city{Berkeley}
  \state{CA}
  \country{USA}
}

\renewcommand{\shortauthors}{Smith et al.}
\acmArticleType{Research}
\acmCodeLink{https://github.com/borisveytsman/acmart}
\acmDataLink{htps://zenodo.org/link}
\acmContributions{}
\begin{abstract}
Large language models (LLMs) generate text that reinforces standard language ideology: a bias towards certain language varieties that are granted more prestige, authority, and legitimacy than others. This paper contributes a sociotechnically grounded faceted taxonomy that illustrates how generative AI systems reproduce standard language ideology and its societal implications. We introduce the concept of \textit{standard AI-generated language ideology} to explain how AI systems confer legitimacy on certain language varieties while marginalizing others, structuring patterns of performance disparity, stereotyping, appropriation, and erasure. We then discuss ongoing tensions around what constitutes desirable system behavior, as well as advantages and drawbacks of generative AI tools attempting or refusing to imitate different language varieties. To address the power relations shaping generative AI and the mechanisms identified in our taxonomy---legitimation, stereotyping, appropriation, and erasure---we offer recommendations that emphasize accountability, community agency, control, and ownership. These recommendations recognize linguistic diversity as a resource to be protected, valued, and sustained as part of a just AI future.
\end{abstract}

\ccsdesc[500]{Computing methodologies~Artificial intelligence}
\ccsdesc[500]{Social and professional topics~Socio-technical systems}

\keywords{large language models, generative AI, algorithmic bias, linguistic justice}

\maketitle

\section{Introduction}

Tools built on large language models (LLMs), such as ChatGPT, promise immense benefits to their users; however, those benefits are not distributed equally across speakers of different language varieties. Rather, ChatGPT---and other popular generative AI language technologies---reproduce and normalize standard language ideology, reinforcing a hierarchy between language varieties and positioning "standard" varieties (particularly "standard" American English (SAE)) as the most appropriate or legitimate form of communication. This is not merely a performance problem, but a sociotechnical one, in which AI systems participate in reproducing linguistic hierarchies with social, cultural, and political consequences. In this paper, we examine how these hierarchies manifest in AI-generated language and present a novel taxonomy of open problems regarding standard language ideology in AI-generated language. Although concurrent work has surveyed challenges in natural language processing (NLP) for different languages and dialects \cite{Joshi2025}, this is among the first to explicitly connect standard language ideology to the behavior of LLMs and the power structures they reproduce, contributing to ongoing FAccT scholarship that frames fairness and harm as sociotechnical rather than purely technical phenomena.

The taxonomy serves two primary purposes. First, it provides a structured lens for analyzing how linguistic biases in AI-generated language reflect and reinforce broader systems of power. This helps researchers identify the language ideologies shaping AI development and aligns with FAccT’s sociotechnical turn toward institutional, political, and cultural determinants of AI behavior. Second, it functions as a diagnostic tool by linking those structural insights to concrete patterns in AI behavior. Together, these functions explain \textit{why} these biases arise and \textit{how} they materialize in practice. Grounded in sociolinguistic theory, our analysis connects documented harms in AI-generated language to long-standing research on language ideology. This connection allows us to show how well-documented hierarchies in language---including beliefs around which languages are considered standard or "better" and linguistic discrimination for speakers of minoritized varieties---are systematically reproduced in generative AI. Situating these dynamics within standard language ideology demonstrates that such issues are not merely technical problems to be solved through engineering fixes, but structural problems that demand critical engagement with power and ideology. Accordingly, our recommendations aim not to mitigate particular harms, but to challenge and reshape the conditions that produce them---a task made especially urgent as governments pursue “sovereign AI” initiatives that may counter English dominance while still reproducing internal linguistic hierarchies.

Our primary contribution is an original faceted taxonomy that operationalizes the novel concept of standard AI-generated language ideology into a diagnostic tool. This taxonomy moves beyond a survey of model errors to categorize the structural mechanisms---legitimation, stereotyping, appropriation, and erasure---through which LLMs reproduce and entrench linguistic power hierarchies.

The paper proceeds as follows. We ground our taxonomy in a  literature review  of standard language ideology, linguistic bias, and linguistic discrimination in LLMs (Section \ref{sec:related-work}). Our methodology (Section \ref{sec:methodology}) leverages our multidisciplinary team of social scientists, computer scientists and linguists to ensure our claims are analytically rigorous and grounded in established theory. We then introduce our taxonomy, which categorizes key open problems in this space and highlights their societal implications (Section \ref{sec:taxonomy}). Finally, we offer recommendations (Section \ref{sec:recommendations}), followed by a discussion of limitations and directions for future work (Section \ref{limitations}), before concluding (Section \ref{conclusion}).

\section{Background}
\label{sec:related-work}
This section first introduces standard language ideology, then reviews evidence of how it appears in AI-generated language. We next outline the sociolinguistic mechanisms that provide the theoretical grounding for our taxonomy, before situating these issues within the broader context of multilingual and “sovereign AI” initiatives.

\subsection{Standard language ideology}
We draw on sociolinguistic and anthropological traditions to understand language ideologies as socially shared beliefs about language and its speakers, deeply intertwined with power, identity, and social hierarchy \citep{WoolardSchieffelin1994, IrvineGal2000, Kroskrity2004}. Language ideologies shape which ways of speaking are perceived as legitimate, appropriate, or valuable, and which are dismissed as incorrect, deficient, or inferior. Standard language ideology is one such ideology: an imposed ``bias toward an abstracted, idealized, and homogeneous spoken language'' whose goal is the ``suppression of variation'' \cite{lippigreen1994}. In other words, it is a common and false belief that some language varieties---usually those used by communities with more social prestige---are ``better'' or ``more complex'' than others. As discussed by \cite{IrvineGal2000}, language ideologies involve multiple intertwined processes. First is \textit{iconization}, by which linguistic features are taken to not only index social features, but to depict the social group's essence: for example, speakers of Standard Quechua may be perceived as more educated but less inherently indigenous than speakers of local Quechua varieties \cite{andronis2003iconization}. Language ideologies also involve \textit{fractal recursivity}, recurrent projection of oppositions between groups to further divide the subgroups: for example, in Mexico, ideologies that position Spanish speakers as modern, urban, and non-indigenous, and Nahuatl (Mexicano) speakers as backwards, rural, and indigenous, are further reproduced among Nahuatl speakers to create similar splits between more urban and rural Nahuatl speakers \cite{messing2002fractal}. Finally, language ideologies make use of erasure, obscuring facts that do not match the ideological narrative: for example, ignoring the consistent and complex grammars of African-American English or Hawai'i Creole English to position them as ``broken English'' \cite{SIEGEL2006157}.

Linguistically, however, all language varieties are equally valid. The idea that certain languages are ``better'' than others ignores the fact that all language varieties are equally capable of expression \citep{Hill2008}. In fact, ``standard'' language is not spoken by any real community, but is an abstracted variety that can only be defined in contrast to the speech of marginalized communities.

Despite this, certain language varieties have been institutionally privileged as more ``standard'' and viewed as more ``appropriate'' than others. This privileged status stems from the language variety's association with people in power. As economic globalization has cast English as a \textit{lingua franca}, or the common language adopted by speakers of different languages \citep{Michaud2015EnglishAA}, discrimination against different dialects of English has become intertwined with questions of discrimination against people of different nationalities and ethnicities \citep{Crystal2003, Phillipson2010-nx, tietze2004spreading}. The privileged position of English is a result of the social and historical power that English speakers have had throughout history, especially as it relates to colonization across many parts of the world. The spread of English as the default, and the increasing imposition of its standard varieties worldwide, has been framed as ``linguistic imperialism'' that threatens other languages and language varieties \citep{Jenkins2013-kc, pennycook2004}.

Research on linguistic discrimination details numerous cases of discrimination against speakers of non-standard varieties, illustrating that standard language ideology operates not only between different languages (e.g., English vs. Spanish), but also within languages, producing internal hierarchies. In the United States, ``Standard'' American English (SAE) functions as the dominant language variety and reflects an abstracted collection of linguistic norms associated with middle-class, white men who have held disproportionate social and political power within the country \citep{Baker-Bell2020-hf, Hill2008, Rosa_Flores_2017}. Other language varieties have been devalued through processes of institutional subordination \citep{Baker-Bell2020-hf, Hill2008, lippigreen1994}. African American English (AAE) is one such language variety, among many others (e.g. Irish English, \cite{moore2024}; Indian English, \cite{kutlu2020negative}; Chicano English, \cite{guerrero2014you}). For example, using AAE has been linked to being denied housing due to ``sounding Black'' \citep{Makoni2002-wb, Wright2023}. Even when using "standard" varieties such as SAE, marginalized people can be subjected to linguistic bias \cite{samy2007critical}. Though linguistic discrimination in English is most well-studied, similar discrimination occurs with varieties of other languages, such as Spanish \cite{pascual, Romn2023} and German \cite{Adler2019}. Because language itself is linked to identity and power \citep{crenshaw1995critical}, language ideologies shape both social and linguistic structures \cite{Woolard2020} in ways that can reinforce social identities and existing power imbalances.

By misleading people to believe that some languages and language varieties are better than others, standard language ideology can perpetuate harmful patterns of linguistic discrimination and the oppression of speakers of ``non-standard'' varieties. Linguistic discrimination can often serve as a proxy for other forms of discrimination along lines of race, gender, nationality, or class, among others \citep{Drodowicz2024}. This discrimination resulting from the promotion of language hierarchy can be subtle under perceptions of benevolence, such as encouraging others to speak more ``appropriately,'' or more obvious bigotry, such as associating certain ways of speaking with lacking intelligence \citep{baugh2005linguistic}. Relatedly, standard language expectations dictate access to social capital through means such as education, employment, or public office \citep{Wiley1996}. Those who speak more closely to the standard varieties benefit from better access to such resources and opportunities.

\subsection{Standard language ideology in large language models}
\label{sec:LLMs}
Existing research demonstrates that standard language ideology manifests in LLMs across multiple dimensions. Empirically, this work has primarily clustered around two areas of concern: (1) performance disparities across language varieties and (2) biased or stereotyped representations of minoritized language varieties. Research on language models preceding the release of ChatGPT has highlighted ways in which these models perform worse for speakers of certain English language varieties. Several studies examine performance discrepancies related to AAE. For example, language models perform worse for AAE on tasks including text generation, sentiment analysis, and parsing \citep{Blodgett2018, field-etal-2021-survey, Groenwold2020}. \cite{thylstrupDetectingDirtToxicity2020} discuss how content moderation tools fail to capture the semantic richness of AAE, such as by making blanket assumptions about the complex semantics of reclaimed slurs. \citep{Lin2024OneLM, ziems-etal-2022-value}
measure performance on different NLP tasks for queries rewritten in AAE and find that tested models' performance worsens on the rewritten queries. Furthermore, \cite{deas-etal-2023-evaluation} discuss how language models including GPT-3, ChatGPT, and GPT-4 have higher perplexity for AAE, interpreted as greater difficulty in understanding AAE, and have trouble producing natural-sounding and semantically accurate AAE text. Outside of AAE, \cite{yong-etal-2023-prompting} find inconsistent performance for ChatGPT's production of Singlish\footnote{Singlish is an English-based creole spoken in Singapore.} and code-switched text involving Southeast Asian languages, finding that even when generated text was perceived as natural (as though produced by a native speaker), it sometimes contained ``semantic inaccuracies…discernible by native speakers.'' These findings are echoed in research demonstrating that LLMs often perform worse for non-English, regional, and traditionally "low-resource" language varieties. 
While this performance gap is often driven by a lack of available training data, data scarcity itself can be understood as a material consequence of standard language ideology. Institutional decisions regarding which languages are prioritized for digitization and optimization reflect underlying hierarchies of value,
revealing systemic inequities rather than isolated technical limitations \citep{Bella2024LanguageBiasFAccT, prinos2024accentfacct, ramesh-etal-2023-fairness}.

In addition to issues of poorer performance for minoritized language communities, language models can  advance stereotypes regarding speakers of a particular language variety \citep{hofmann2024dialect, nee2022}. \cite{finch2025findingvoiceevaluatingafrican} note issues in language models' production of AAE, finding that AAE speakers prefer chatbots speaking "mainstream" or "standard" English varieties over AAE-speaking ones. \cite{hofmann2024dialect} illustrate that language models produce text with harmful stereotypes about speakers of AAE through ``dialect prejudice." 

At a higher level, illustrating the hierarchy of English language varieties in LLMs, a study comparing eight non-standard varieties of English to Standard American and British English found that language model outputs tend to default to standard varieties, particularly SAE, even in response to non-standard varieties \cite{fleisig2024}. 
Responses to the eight non-standard varieties all had over 84\% reduction in linguistic features from that variety, with over 96\% reduction for Jamaican, Singaporean, Scottish, and African-American English. Meanwhile, outputs default to SAE orthography. When native speakers of each variety reviewed the AI-generated outputs, outputs to non-"standard" varieties consistently exhibited issues including lack of comprehension (10\% worse compared to "standard" varieties) and increased stereotyping (16\% worse), condescension (12\% worse), and demeaning content (22\% worse). Together, these studies show that LLMs do not merely “struggle” with non-standard varieties. Rather, their outputs actively normalize SAE as the linguistic default, perpetuate linguistic stigmas, and reinforce social hierarchies.

Performance discrepancies and stereotypes present in LLMs are linked to the language data that underpins these technologies. English disproportionately dominates the datasets that power LLMs, a consequence of LLMs relying on language data from the Internet, particularly from Common Crawl \cite{baack2024critical}. Internet data overwhelmingly represents English, and Internet use varies due to social factors. An estimated 60\% of all language content on the Internet is in English, despite only ~17\% of people speaking English globally \citep{nee2022}. Meanwhile, 88\% of languages have ``exceptionally limited resources'' in digital spaces \citep{joshi-etal-2020-state}. Non-standard varieties of English are disproportionately removed from scraped Internet data when scraped pages are filtered using blocklists \cite{dodge-etal-2021-documenting}. Even if languages are well-represented in digital corpora, certain perspectives are over- or under- represented. For example, on Reddit, users are 67\% male and 70\% white, resulting in potential reinforcement of white, male perspectives (\citep{bender2019benderrule} citing \citep{barthel2016reddit}). Other perspectives may be actively targeted, cyberbullied, and otherwise marginalized online. Research exposes growing concerns of online gender-based harassment with clear links between online spaces and misogyny, which can be further amplified for Black and brown women \citep{ging2018, duggan2017online, rights2014violence}. This harassment and the resulting safety and mental health implications can result in decreased use of online spaces. Meanwhile, although Black people have historically been overrepresented on Twitter (now X) compared to other demographics in the general US population \citep{brock2012}, Black Tweets are still often considered ``inappropriate'' and are more often inaccurately flagged as hateful by automatic hate speech detection tools \citep{davidson-etal-2019-racial}. Beyond being technically inaccurate, this disproportionately censors Black speakers and shapes what, and whose, language exists in training corpora. Crucially, these are not merely issues of uneven data availability; they are mechanisms through which certain varieties come to be constructed as standard. Decisions about whose language is collected, filtered, amplified, or removed shape which forms of speech AI systems learn to treat as normal, legitimate, and default.

Research on language models treats English, particularly SAE, as the status quo, with work on other languages ``often considered `language specific' and thus reviewed as less important than equivalent work on English'' \citep{bender2019benderrule}. This default of SAE partly reflects the people who hold disproportionate power in these spaces. Leading industry developers of LLMs, such as Open AI, Google, Microsoft, and Meta, are largely headquartered in the United States, with employees and corporate leadership that skew white and male \citep{omara2022why}. Demographics of AI researchers in industry and academia are likewise skewed \citep{west2019discriminating}, with NLP researchers from Latin America, Africa, the Middle East, and Southeast Asia particularly underrepresented \citep{rungta-etal-2022-geographic}. While the demographics of those developing and leading NLP research and companies do not directly result in the reinforcement of SAE at the level of language models, this lack of diversity in the technologies' developers and managers can lead to similar thinking in ways that miss potential biases the technology can have \citep{cowgill2020biased} with a default towards an (inequitable) status quo. 

Taken together, research illustrates the ways that standard language ideology is reflected in LLMs and AI-generated language, including through poorer performance and greater stereotyping for minoritized language varieties. Meanwhile, "standard" English varieties, and particularly SAE, remains the default output. These findings align with broader research arguing that conversational AI systems materialize specific language ideologies rather than neutral linguistic competence, codifying notions of “proper,” “civil,” and “standard” language \citep{hohn2024languageideology, chronis2025nlpideology}. These are not accidental results of data imbalance or engineering limitations; they reflect sociotechnical choices about whose language is preserved, filtered, amplified, or removed. 

\subsection{Mechanisms of standard language ideology: Theoretical grounding for our taxonomy}
\label{sec:mechanisms}
Sociolinguistic scholarship has shown that standard language ideology does not simply privilege certain language varieties; it operates through recognizable social and institutional processes. It establishes some language forms as legitimate and authoritative with others as deficient or inappropriate \citep{Lippi-Green2012-ex, MilroyMilroy1999, Bourdieu1991}. These hierarchies are sustained through racialization and stereotyping, whereby linguistic forms become morally and socially typified and tied to marginalized identities \citep{Agha2012, Eckert2012, RosaFlores2017}. Dominant institutions also appropriate and commodify minoritized language practices, extracting cultural and social value while separating them from the communities that produce them \citep{Hill1998, HellerDuchene2012}. Finally, institutions frequently silence or render invisible linguistic forms deemed illegitimate through processes of erasure \citep{IrvineGal2000}. 

Together, these four mechanisms---legitimation, stereotyping, appropriation, and erasure---provide the theoretical grounding for our taxonomy. They explain how standard language ideology produces and sustains linguistic inequality. In Section \ref{sec:taxonomy}, we examine how generative AI systems reproduce these dynamics in computationally mediated form.

\subsection{Sovereign AI and the politics of language representation in generative AI}
These dynamics are particularly salient in the context of emerging “sovereign AI” initiatives, where states seek to develop domestic AI capacity to reduce reliance on corporate platforms and assert technological autonomy \citep{Vogiatzoglou2025DigitalSovereignty, Floridi2020DigitalSovereignty}. Multilingual national AI efforts such as India’s BharatGen and broader IndiaAI strategy explicitly position language inclusion as part of this sovereignty agenda, emphasizing support for multiple Indian languages and cultural bias mitigation \citep{indiaBharatGen, IndiaAI2023Strategy}. Such initiatives appear to counter standard language ideology by expanding linguistic inclusion. However, sovereign AI does not automatically dissolve linguistic hierarchy; instead, it relocates its center \citep{HellerDuchene2012, Mohamed2020DecolonialAI}. Decisions about which languages are prioritized, which varieties are treated as “standard”, and whose speech is recognized as legitimate become matters of institutional design and political authority. In this sense, sovereign and multilingual AI projects may challenge global English dominance while still risking the reproduction of internal language hierarchies, underscoring why interrogating standard language ideology in LLMs remains both urgent and unresolved.

\section{Methodology}
\label{sec:methodology}
The taxonomy is grounded in a conceptual framework that synthesizes empirical evidence with insights from linguistic theory, sociolinguistics, and applied linguistics. It identifies and organizes a set of open problems that illustrate how standard language ideology is reproduced in generative AI outputs, alongside their broader societal implications. To identify these open problems, we conducted a focused review of large-scale studies and systematic evaluations examining linguistic bias and discrimination in LLM outputs, complemented by a broader literature review on standard language ideology, linguistic inequality, and language-based harm. This linguistic and theoretical grounding is essential for understanding the broader and longer-term societal implications of language technologies. While LLMs are a recent technological development, the ideological forces they reproduce have been extensively studied in prior research on language, power, and inequality. Our taxonomy builds on this body of work to surface how those same dynamics are rearticulated in AI-mediated contexts.

Our analytic process was both deductive and inductive. Guided by sociolinguistic theory, we began with established mechanisms through which standard language ideology operates (e.g., legitimation, racialization and stereotyping, appropriation and commodification, and erasure, as outlined in Section \ref{sec:mechanisms}) and used these as an initial conceptual scaffold. We then iteratively mapped empirical findings from existing studies onto these categories, refining them in response to recurring patterns in the data. Where empirical patterns did not neatly align with initial theoretical expectations, we refined our analytic framing to better reflect patterns documented in the empirical research.

We categorized studies according to the kinds of harms they documented and the linguistic dynamics they revealed, and for each category we synthesized documented AI manifestations and broader societal implications. The “documented AI manifestations” column in our taxonomy synthesizes empirically demonstrated system behaviors. The “potential societal implications” column extends these findings by drawing on sociolinguistic theory and prior scholarship on language-based inequality to articulate likely downstream consequences. We acknowledge that these implications are not exhaustive nor universally applicable; rather, they articulate plausible trajectories rooted in existing evidence and theory. Through this iterative process, we arrived at a taxonomy designed to be conceptually rigorous, empirically grounded, and analytically useful. By situating the taxonomy at the intersection of empirical studies of LLM behavior and critical sociolinguistic scholarship, this paper offers a structured framework for understanding and addressing the reproduction of linguistic hierarchies in generative AI systems. 

\section{Taxonomy of open problems regarding standard language ideology in AI-generated language}
\label{sec:taxonomy}
We introduce a faceted taxonomy of open problems arising from reproduction of standard language ideology in AI-generated language technologies, alongside implications for different language communities globally (see Table 1). Existing research has primarily examined these dynamics through two dominant empirical lenses: performance disparities across language varieties and biased or stereotyped representations of minoritized varieties. These represent only part of the broader ideological landscape. When understood through sociolinguistic theory, these patterns do not appear as isolated failures, but as expressions of deeper mechanisms through which standard language ideology traditionally operates. Our taxonomy synthesizes the empirical findings discussed in Section \ref{sec:LLMs} with theoretical insights to identify five open problems that more fully characterize how AI systems can differentially legitimize, stereotype, appropriate, or erase minoritized language varieties. In the sections that follow, we elaborate on each open problem, illustrating how it manifests in current systems and how it contributes to the reproduction of linguistic hierarchy in technologically mediated contexts.


\begin{table*}[t]
\centering
\caption{Taxonomy of Open Problems in AI Reproduction of Standard Language Ideology}
\footnotesize
\setlength{\tabcolsep}{4pt}
\begin{tabular}{
  >{\raggedright\arraybackslash}p{2.7cm}
  >{\raggedright\arraybackslash}p{3.6cm}
  >{\raggedright\arraybackslash}p{4.2cm}
  >{\raggedright\arraybackslash}p{4.2cm}
}
\toprule
\textbf{Facet / Open Problem} 
& \textbf{Connection to Standard Language Ideology} 
& \textbf{Documented AI Manifestations} 
& \textbf{Societal Implications} \\
\midrule

\textbf{Normalization of ``standard'' language as default}
& Reinforces and legitimizes ``standard'' varieties— particularly SAE—as neutral, authoritative, and universally appropriate.
& \begin{itemize}[leftmargin=*]
  \item Defaults to SAE even when non-standard varieties are present \citep{fleisig2024, hofmann2024dialect}
  \item Removes dialectal features in paraphrasing \citep{fleisig2024}
  \item Automation bias may reinforce these norms \citep{goddard2012automation, horowitz2024}
\end{itemize}
& \begin{itemize}[leftmargin=*]
  \item Naturalizes linguistic hierarchy
  \item Legitimizes prestige norms
  \item Encourages conformity and internalized bias
\end{itemize} \\
\midrule

\textbf{Unequal quality of service for minoritized varieties}
& Reflects historical treatment of some language forms as less legitimate or less worthy of institutional support.
& \begin{itemize}[leftmargin=*]
  \item Lower comprehension / degraded responses \citep{Blodgett2018, Groenwold2020, Lin2024OneLM, deas-etal-2023-evaluation}
  \item Misclassification as hateful/toxic → refusals \citep{davidson-etal-2019-racial}
  \item Users face greater difficulty than ``standard'' speakers \citep{crawford2017, mccluney2019}
\end{itemize}
& \begin{itemize}[leftmargin=*]
  \item Unequal access to AI tools
  \item Defines who is a ``proper'' technology user
  \item Incentivizes digital code-switching with psychological tolls
\end{itemize} \\
\midrule

\textbf{Stereotyped and racialized production}
& Mirrors practices of marking minoritized varieties as deviant, humorous, uneducated, or morally suspect.
& \begin{itemize}[leftmargin=*]
  \item Stylized or caricatured production \citep{hofmann2024dialect, nee2022}
  \item Disproportionate demeaning or stigmatizing content \citep{fleisig2024}
  \item Native speakers rate outputs as more demeaning \citep{fleisig2024}
\end{itemize}
& \begin{itemize}[leftmargin=*]
  \item Normalizes prejudice as ``data-driven truth''
  \item Reinforces negative public perception
  \item Harms dignity and cultural recognition
\end{itemize} \\
\midrule

\textbf{Appropriation and commodification}
& Continues extraction of cultural and linguistic value from marginalized communities without accountability.
& \begin{itemize}[leftmargin=*]
  \item Enables outsiders to appropriate identity styles \citep{Bucholtz2011, eberhardt2015, MatamorosFernndez2022, sobande2021, Holliday2023, feffer2023deepdrake}
  \item Circulates language without attribution or context \citep{chery2022black, finch2025findingvoiceevaluatingafrican}
  \item Facilitates impersonation / linguistic minstrelsy \citep{jerkins2022}
\end{itemize}
& \begin{itemize}[leftmargin=*]
  \item Cultural exploitation, surveillance, and extraction
  \item Redistribution of value upward
  \item Erodes trust and authenticity
\end{itemize} \\
\midrule

\textbf{Suppression and erasure}
& Aligns with institutions that regulate ``proper’’ communication and erase forms deemed illegitimate or risky.
& \begin{itemize}[leftmargin=*]
  \item Refusal to generate responses \citep{davidson-etal-2019-racial}
  \item Creates pressure to convert to ``standard’’ forms that receive higher quality responses \citep{fleisig2024}
  \item Disproportionate moderation \citep{davidson-etal-2019-racial}
\end{itemize}
& \begin{itemize}[leftmargin=*]
  \item Limits communicative agency
  \item Narrows linguistic futures in AI ecosystems
  \item Reinforces institutional erasure
\end{itemize} \\
\bottomrule
\end{tabular}
\vspace{4pt}
{\footnotesize *Documented manifestations are discussed further in Section \ref{sec:taxonomy}.\par}
\end{table*}

\textbf{Normalization of "standard" language as default.}
At a high level, generative AI technologies default to ``standard'' language varieties in their outputs, particularly SAE. As discussed in Section \ref{sec:LLMs}, this pattern has been documented in empirical analyses of LLM outputs, showing systematic defaulting to prestige English varieties and substantial reduction of features from non-standard varieties \citep{fleisig2024, hofmann2024dialect}. This reflects the broader mechanism of legitimation, wherein certain forms of speech are subtly reinforced as ``correct'' or ``more appropriate'' way of communicating \citep{Lippi-Green2012-ex, MilroyMilroy1999, Bourdieu1991}. 

This can have a cascading effect that impacts people's own linguistic biases and perceptions of language hierarchies. Lessons can be learned from the dominant position that SAE has held in educational settings in the United States. Despite the prioritization of SAE in classrooms often being seen by teachers as a means of unifying students through a common language \citep{delpit1992, Smitherman1986-fd}, it has negative implications for minoritized students. For example, Black students who speak AAE feel compelled to adopt SAE and dominant ideologies, which then impact their identity expression, academic achievement, and self-perception \cite{lyn2022}. Additionally, due to automation bias (a psychological tendency to over-rely on automation that can result in complacency regarding automation outputs \citep{goddard2012automation}) humans can have greater trust and overconfidence in AI \cite{horowitz2024}. If LLM outputs that prioritize SAE or disparage other dialects are seen as particularly trustworthy, they could exacerbate the idea that ``standard'' languages are more ``correct'' ways of speaking. 

\textbf{Unequal quality of service for minoritized varieties.}
Consistent with prior work demonstrating degraded performance and misclassification of minoritized varieties, particularly AAE, across NLP and LLM systems \citep{Blodgett2018, Groenwold2020, Lin2024OneLM, deas-etal-2023-evaluation}, these disparities constitute not simply technical shortcomings but allocational harms. At a high level, these tools may fail to comprehend the user's prompt, and result in an incorrect or unhelpful response. Relatedly, generative AI tools can more often misclassify inputs of minoritized language varieties as hateful or offensive speech, resulting in refusal to respond \citep{davidson-etal-2019-racial}. This reflects long-standing patterns in which some language varieties are deemed less legitimate and therefore receive reduced institutional support \citep{Lippi-Green2012-ex, MilroyMilroy1999, Bourdieu1991}. 

These behaviors result in speakers of minoritized varieties experiencing more difficulty in using language models than speakers of ``standard'' varieties, which is a form of allocational harm \cite{crawford2017}. Furthermore, if users of generative AI tools are aware that using a minoritized variety results in poorer performance, they may be incentivized to use a ``standard'' variety instead, which can reinforce the stigmatization of non-standard varieties and result in ``digital code switching.'' Code switching, which results from a pressure minoritized speakers can feel to conform to more ``standard'' language, causes psychological tolls \citep{mccluney2019}. These psychological tolls could extend to the digital sphere, though additional research is needed. 

\textbf{Stereotyped and racialized productions of minoritized varieties.} 
Generative AI technologies may be prompted by the user to produce minoritized language varieties, or may be designed to respond to the particular language variety of the input. This may not always be a source of harm, particularly if users of that particular variety are asking for responses back in their language variety. However, responses in minoritized varieties can disproportionately carry or convey stereotypes or demeaning content. Empirical studies demonstrate this dynamic across both pre-ChatGPT NLP systems and contemporary LLMs, which reproduce dialect prejudice, racialized linguistic caricatures, and disproportionate production of demeaning content for minoritized varieties \citep{hofmann2024dialect, fleisig2024, nee2022}. In response to inputs of minoritized languages, models prompted to imitate the input variety may produce a stereotyped version of that language variety that does not accurately reflect the range of linguistic features used by that community. 
This behavior perpetuates the association between speaking a non-standard variety and stereotypical traits of that speaker community, which often serves as a covert form of racism, xenophobia, or other widespread harms (\citep{rickford2016language}; see also Section \ref{sec:LLMs}). The outputs could also convey demeaning content related to speakers of that language community. Examining ChatGPT outputs that imitated the language variety of the inputs, \cite{fleisig2024} found that native speakers of minoritized varieties rated outputs as more often conveying demeaning content than responses to standard English varieties. This reflects the mechanism of racialization and stereotyping, where linguistic form becomes a proxy for social identity and moral judgment \citep{Agha2012, Eckert2012, RosaFlores2017}.

\textbf{Appropriation and commodification of minoritized language practices.}
Even if a language model correctly produces text that reflects the grammar of that minoritized variety, there may be concerns over appropriation (particularly if non-native speakers of that variety are asking for and using this text). Because speaking a non-standard variety can carry covert prestige (i.e., social value associated with use of canonically minoritized language varieties \citep{labov2006social}), use of non-standard varieties by non-native speakers, particularly white speakers appropriating AAE, has been discussed as ``linguistic minstrelsy'' or ``figurative blackface'' \citep{Bucholtz2011, eberhardt2015}. Parallel concerns have been raised regarding the use of speech, music, and image generation tools to imitate people of color, sometimes called ``digital blackface'' \citep{MatamorosFernndez2022, sobande2021, Holliday2023, feffer2023deepdrake}. As LLMs improve at mimicking minoritized varieties, these models may increasingly contribute to similar harms. This aligns with the broader mechanism of appropriation and commodification: models enable the extraction of cultural and linguistic value from marginalized communities without accountability or benefit to those communities \citep{finch2025findingvoiceevaluatingafrican, MatamorosFernndez2022}.
This extraction facilitates a redistribution of value upward, as the cultural and social capital of minoritized language is transformed into commercial assets for dominant technology providers. This process erodes trust and authenticity by removing community agency over how their linguistic heritage is represented, effectively alienating speakers from their own varieties as they become commodified by outsiders.

In addition, use of uncredited language from marginalized groups by language models can further cultural appropriation and erase the linguistic history of speaker communities. A popular ``Gen Z Translator'' extension of ChatGPT\footnote{https://chat.openai.com/g/g-AbhjZGbhY-gen-z-translator} that claims to ``transform any short text to be Gen Z slang filled'' frequently produces AAE. However, rebranding expressions that emerged in Black communities as generational vernacular, as opposed to vernacular linked to ethnic and racial communities, represents an institutional erasure where dominant tech developers render the context and historical origins of language invisible. It risks ``appropriating Black culture and perpetuating racism as [speakers] take on Black speech without assuming Black Americans' struggle'' \cite{chery2022black}.

Producing minoritized varieties could also result in manipulation of those language communities by people outside of them. Readily available ways of producing text that imitates a non-standard variety could help agents to feign in-group membership for malicious purposes, such as the case of Russian misinformation bots posing as Black people online \cite{jerkins2022}. 
More broadly, bots capable of understanding and imitating minoritized language varieties could facilitate surveillance of those communities by enabling infiltration of community spaces under the guise of linguistic membership.

\textbf{Suppression and erasure of minoritized varieties.}
Preventing language models from producing anything but prestige or ``standard'' varieties of a language avoids the potential harms discussed regarding appropriation and manipulation. However, if speakers of a marginalized variety want the model to reply in that variety, then it may constitute a quality of service harm because, unlike speakers of the ``standard'' variety, they are unable to fully interact with the model in their native variety. This can also contribute to erasure of minoritized languages. In practice, attempts to “avoid harm” by suppressing non-standard output risk reproducing long-standing patterns of linguistic erasure, echoing broader findings that automated systems disproportionately censor, misclassify, or delegitimize minority language practices \citep{davidson-etal-2019-racial}. This reflects the mechanism of erasure, wherein linguistic forms deemed illegitimate are pushed out of social and technological infrastructures \citep{IrvineGal2000}, further reinforcing language hierarchies and the false notion of “correct” ways of speaking.

\subsection{Standard AI-generated language ideology}
The mechanisms detailed in our taxonomy coalesce into what we term standard AI-generated language ideology. We introduce this as a novel concept that formalizes how certain language varieties, particularly SAE, are positioned as the normative, appropriate, and default mode of linguistic production, while marking other varieties as deviant, risky, or disposable. Default production of “standard” varieties reinforces the notion that there are correct or superior ways of communicating, with psychological and social implications for speakers of minoritized varieties who encounter their language treated as less legitimate. This ideology is further sustained through unequal quality of service: when LLMs perform worse for minoritized language varieties, the system creates a functional incentive for users to “digitally code switch” into standard varieties to access functionality, legitimacy, or institutional service. 
Meanwhile, when models do attempt to produce minoritized varieties, they disproportionately do so in stereotyped, racialized, appropriative, or otherwise harmful ways, reinforcing stigmatizing associations rather than linguistic diversity. Consequently, these dynamics move beyond technical performance; they shape who is afforded linguistic belonging within AI systems and who must adapt, assimilate, or risk harm. 

As generative AI tools increasingly mediate everyday activities---from accessing public services to educational and health resources---these systems risk naturalizing linguistic hierarchy while amplifying Western, and particularly American, linguistic dominance. In this sense, AI-generated language does not merely reflect existing power asymmetries; it helps reproduce and legitimize them as technologically neutral, and in some cases may also reshape how they are enacted in technological infrastructures. This ideology is not limited to corporate or U.S.-centric AI systems. 
As sovereign and multilingual AI initiatives expand, the question is not only whether English remains dominant, but which languages and varieties are elevated as “standard” within national and regional infrastructures. In this sense, sovereign AI may redistribute linguistic authority without necessarily dismantling hierarchy, reinforcing the importance of interrogating how standardness is constructed in AI systems.

\subsection{No singular "correct" behavior for language models}
The issues outlined above raise a fundamental question: what should count as desirable or appropriate behavior for generative language technologies? Different design choices inevitably privilege different speakers and produce different kinds of harm. Our taxonomy highlights the difficulty technologists face in determining what “fair” or “responsible” behavior looks like in practice. Central to these dilemmas is the fact that language is not simply a communication tool, but a resource to reflect identity, affiliation and social positioning. For example, speakers may switch to a non-standard variety when speaking to someone to signal their shared membership in an in-group, indexing solidarity and belonging \citep{auer2013, gumperz1982discourse}. But generative AI systems do not have social identities in the same way people do. Should a model reflect the positionality of its creators (often a white, Western, and male-dominated group)? Should it aspire to a “view from nowhere,” if such neutrality is even possible? Should expectations vary depending on who trained the model, whose data it relies on, and who benefits from it? These are not technical questions alone; they are normative decisions about linguistic authority that are typically made by model creators and tech industry members, but they fundamentally depend on the nuanced norms of speaker communities that rarely have a say in how these models are designed. 

Historically, advocates of linguistic standardization have argued that privileging a “standard” variety can promote mutual intelligibility, administrative coordination, and educational access \citep{MilroyMilroy1999, Lippi-Green2012-ex}. From this perspective, defaulting to a standardized variety in AI systems might appear desirable or pragmatic. Yet such benefits have never been evenly distributed, and they have often come at the cost of stigmatizing or marginalizing non-standard speakers. Moreover, contemporary AI technologies are theoretically capable of supporting linguistic plurality---through translation, dialect-sensitive interaction, and community-specific tooling---without requiring universal convergence to a single norm. 

For popular generative AI chatbots and voice tools developed and owned by large corporations (e.g., ChatGPT, Gemini, Voice Engine) there is no clear ``correct'' behavior. Automatically defaulting to “standard” varieties like SAE risks reinforcing linguistic hierarchy, stigma, and pressures toward digital code switching, echoing documented harms in education and other institutional contexts \citep{lyn2022}. While automatically mimicking minoritized language varieties risks stereotyping, appropriation, and manipulation of those communities, as our taxonomy demonstrates. These tensions are not easily resolvable as they reflect longstanding sociolinguistic tradeoffs about legitimacy, belonging, and power rather than simply technical design flaws.

Although it is unclear what approach should be taken for behavior of generative AI language technologies, it is clear that these decisions are currently centralized in a small number of for-profit technology corporations whose incentive structures are oriented toward shareholder value rather than linguistic justice or community accountability. Thus, the question is not merely how models should behave, but who has the authority to decide.

\section{Recommendations: Moving towards emancipatory outcomes}
\label{sec:recommendations}
Instead of asking what the ``appropriate'' way is for generative AI models to behave, we propose a more fundamental question: how might recognizing and confronting standard language ideology guide the development of generative AI toward more emancipatory and socially just outcomes? This reframing is crucial because there is no universally "correct" way for models to respond across all contexts (e.g., there is not a clear answer around what the default production of models \textit{should} be), and a narrow focus on performance can obscure deeper social and political stakes that standard language ideology calls us to examine. Therefore, rather than aligning our recommendations to each individual open problem in the taxonomy, we take a higher-level perspective that addresses the power relations that structure generative AI systems, including accountability, community agency, control, and ownership. We present three concrete actions for AI researchers, funders, and practitioners. These recommendations serve as structural responses to the mechanisms documented in our taxonomy---legitimation, racialization and stereotyping, appropriation and commodification, and erasure. The recommendations therefore target the institutional and epistemic conditions under which standard language ideology becomes encoded into AI systems. 

\subsection{Expand evaluation metrics to include language variety harms.} First, and most simply, the fact that language models perpetuate discrimination on the basis of language variety means that evaluation of language models for potential harms (e.g. ``toxicity'' evaluations) should expand to capture discrimination based on the user's language variety. Measurement of harms due to standard language ideology is particularly important not only to measure bias, but to reshape what counts as meaningful model performance and ensure accountability to different language communities.

We therefore recommend that AI developers and researchers do the following: (a) Incorporate language variety as a specific category in model evaluation frameworks alongside race, gender, and other demographic attributes; (b) develop benchmarks and metrics that assess whether models perpetuate documented and theoretically anticipated harms (e.g., erasure of linguistic features, stereotyping, and condescension) against different minoritized language varieties; and (c) publicly share evaluation results to ensure transparency and accountability, while raising awareness about the risks of reinforcing linguistic discrimination and hierarchies through AI generated language. By incorporating language variety as a category for evaluating performance and "toxicity" and being transparent about results, we can better capture and understand the implications of AI technologies on marginalized language communities globally. 

\subsection{Support speakers of different language communities in creating their own datasets and models.} Issues surrounding appropriate responses to minoritized varieties, and when/if responding in the variety is appropriate, stem in part from the fact that these models are developed by organizations that are not composed of members of these speech communities. Instead, the development of language models by speaker communities, for speaker communities, helps to avoid this top-down imposition of linguistic norms (see, for example, \citep{mahelona2023openai, nekoto-etal-2020-participatory}, and the broader discussion of indigenous data sovereignty by \citep{kukutai2016indigenous}). Importantly, community-led dataset and model development directly challenges the logics of appropriation and erasure that underpin standard language ideology, shifting epistemic authority back to language speakers themselves.

There are various examples of communities whereby speakers of minoritized language varieties are collecting data and developing language models. One example is Masakhane \cite{orife2020masakhane}, a grassroots community focused on NLP research ``in African languages, for Africans, by Africans.'' With over one thousand participants in their community across 30 African countries, they have developed translation results for approximately 40 African languages and supported open-source datasets and subsequent machine translation models to address under-representation of African languages in technology. Another example is AI4Bharat \cite{kunchukuttan2020ai4bharat}, a research lab at IIT Madras that "is dedicated to advancing AI technology for Indian languages through open-source contributions." The organization seeks to build language models and datasets for various Indian languages across India's diverse linguistic landscape.

Some cases go further by ensuring that community members have a say in how their data is used. Te Hiku Media---a Māori-owned and operated nonprofit based in New Zealand---focuses on preserving and promoting the Māori language (te reo Māori) and culture \cite{james2020developing}. In 2018, they launched a competition in which Māori speakers across New Zealand recorded over 300 hours of annotated audio. With this, they built AI tools for automatic speech recognition and pronunciation of te reo Māori. Te Hiku Media created various AI technologies and tools for Māori (e.g., automatic speech recognition, auto tagging, real time language feedback). They also established new data protection protocols and worked with Māori-led organizations to develop AI tools that support the community. Simultaneously, they worked with other Indigenous communities to leverage the Te Hiku language database to train language models for other Indigenous groups. The diversity of methods adopted by these organizations (cross-national collaboration, open-source contributions, and community-centered competitions) illustrates how unconventional methods of research can help to center the needs of non-``standard'' speech communities. In 2024, a data sovereignty agreement between Microsoft and Te Tumu Paeroa (the office of the Māori trustee) was established, in which Te Tumu Paeroa, acting as the anchor tenant for Microsoft’s New Zealand-based data centers, ensured that Māori data would be stored locally and governed according to a Māori data sovereignty framework. This agreement reflects an institutional recognition and honoring of Indigenous data governance that supports community ownership.

We recommend that multilateral institutions, funders, and policymakers prioritize meaningful support---financial, technical, and infrastructural---for language communities' own dataset and AI development efforts. Community-led development helps prevent top-down imposition of linguistic norms. In providing support, ensuring data collection is not extractive while allocating funding for data collection efforts and people whose data is collected remains important. In some cases, communities may choose to practice refusal (opting not to share or use their data in the development of AI technologies by outside actors) or may prioritize data sovereignty above participation. Supporting diverse language communities means supporting these choices.

Just as community-led dataset and model development empowers speakers to shape technologies around their own linguistic practices, "sovereign AI" initiatives can be understood as national-scale efforts to do similar work---seeking to build locally governed AI infrastructures to serve their language communities and nation. Yet these initiatives must not simply replicate standard language ideology at a different institutional scale. As discussed earlier, sovereign AI may expand linguistic inclusion while still recentralizing linguistic authority at the level of the state. We therefore recommend that sovereign AI efforts: (1) embed community governance rather than confining decision-making to state or corporate actors; (2) support dialectal and indigenous language inclusion, not only national "standard" forms; and (3) adopt data sovereignty and benefit-sharing frameworks that materially return value to speaker communities. At the same time, because not every language or variety can be supported at once, prioritization must be explicit, transparent, and accountable. Sovereign AI initiatives should clearly articulate which languages are prioritized, why those choices were made, who participated in decision-making, and how tradeoffs were justified. Making these processes transparent supports conditions for public scrutiny and democratic discussion, rather than silent consolidation of linguistic authority---and the broader political and racial hierarchies it is bound with.

\subsection{Innovate on and implement participatory processes that center marginalized communities, while rethinking language data practices to avoid extractive approaches. } 
While community-led models are essential, the dominance of a small set of LLMs means these companies and developers still bear responsibility to improve model quality for, meaningfully engage with, and redistribute decision-making power to language communities. We recommend two complementary directions.

\textbf{First, embed participatory governance and power-sharing.} AI researchers and practitioners should: (a) engage minoritized language communities as equal partners throughout the AI development lifecycle, from data collection and model training to deployment and evaluation; (b) support community ownership by developing language models by and for speaker communities that reflect the values, norms and linguistic practices of communities they serve; and (c) implement participatory governance structures that give communities the power to make decisions about how their language data is used and tools are deployed. These participatory processes allow community stakeholders to counter the dominance of large tech companies and community outsiders who have, historically, unilaterally determined whose language "counts" in technological infrastructures.

An example is the partnership between OpenAI and the Icelandic government. While most of Iceland's citizens speak English or another second language, its integration with the U.S. and Europe has put the country's native tongue, Icelandic, at risk. With concerns that Icelandic cannot remain the default language in the face of digitalization, the Icelandic government partnered with OpenAI to use GPT-4 in the preservation effort of the Icelandic language \cite{iceland}. Within this, Mi$\eth$eind ehf, a language technology company, assembled a team of 40 volunteers to train GPT-4 on Icelandic grammar and cultural knowledge. Here, AI practitioners (with the help of the community) sought to learn from the speakers to improve model performance. However, OpenAI is benefiting from the engagement of the particular linguistic community and it is unclear to what extent the linguistic community is sharing in the financial benefits over time. 
This illustrates a core tension in addressing allocative harms: while partnerships with large corporations provide the technical infrastructure necessary for high-performance models, they often involve a trade-off where immediate linguistic utility is gained at the expense of equitable value distribution and long-term data control. However, this tension is not inevitable; it persists because current corporate models prioritize data extraction over power-sharing. To resolve this,
the initiative could go further by ensuring that Icelandic speakers are central actors in the development and management of generative AI tools that use and leverage their language and the work they provided to improve the models.

Despite the difficulty of implementing participatory practices in model design, setting ``ambitious, yet reasonable goals'' that help to move from one of the ``extreme poles of transactional consulting and transformative empowerment'' to the other \cite{delgado2023} could help to improve power dynamics involving speakers of minoritized varieties. In addition, participatory AI practices include compensating them fairly for their contributions and respecting indigenous data sovereignty principles. While participatory processes may require additional time and resources, they are critical to create AI systems that empower rather than further marginalize communities. 

\textbf{Second, rethink language data practices to promote non-extractive participation.} 
AI researchers and practitioners must rethink their approach to language data by recognizing what data these models are (and are not) trained on, how they can better learn from minoritized language varieties in non-extractive ways, and how marginalized communities can be enabled as central actors in the development and ownership of generative AI tools. We recommend that AI researchers and practitioners: (a) audit existing language datasets to identify disparities in representation and performance for different language varieties; (b) incorporate diverse language into training data, ideally in ways that build from  recommendation 2; and (c) collaboratively develop mechanisms to compensate and credit communities whose language data is used in model training. As researchers strive to move from merely consulting language communities to empowering them regarding the use and management of their data \cite{delgado2023}, the examples we describe in each of these directions can serve as case studies for future collaboration. 

These recommendations move beyond optimizing model outputs to interrogating and reshaping the institutional, epistemic, and political conditions through which language becomes encoded in AI. Rather than treating linguistic harms as technical anomalies, we frame them as consequences of standard language ideology operating within computational systems. Addressing them therefore requires better measuring harms, redistributing power, centering marginalized language communities, and reimagining what “good” language technology should serve and for whom.

\section{Limitations and Future Research}
\label{limitations}
This paper synthesizes existing empirical work and sociolinguistic theory to develop our faceted taxonomy, grounded in a conceptual framework. As such, the taxonomy is necessarily shaped by the scope and biases of the scholarship it draws upon. While we reviewed a broad body of research, the open problems and implications we identify are not exhaustive. Our taxonomy was developed through a combination of deductive grounding in sociolinguistic theory and inductive synthesis of prior empirical studies, rather than new large-scale model audits or ethnographic fieldwork. As such, it reflects patterns visible in currently documented research landscapes. Future empirical work may add, revise, or contest these categories. 

Second, our analysis is primarily grounded in English, especially the politics of SAE. While we incorporate multilingual and “sovereign AI” contexts, we do not claim that standard language ideology manifests identically across all linguistic, geopolitical, or colonial histories. In some settings, standardization may be community-driven, strategically embraced, or connected to language revitalization rather than domination. Future work should examine how standard language ideology manifests differently across linguistic ecosystems; how national AI projects may both challenge and reproduce internal hierarchies; and how indigenous, regional, and minoritized language communities experience, resist, or reshape these infrastructures. Addressing these questions is timely and will require deeper collaboration with language communities and expanded empirical evaluation across diverse language varieties and communities. 

Finally, there is no single “correct” or universal model behavior with respect to language variety. Communities will disagree about trade-offs. Our contribution is not to resolve these tensions, but to foreground the structural power relations in which they are negotiated. Continued collaboration with language communities and interdisciplinary researchers will be essential to refine and extend this work.

\section{Conclusion}
\label{conclusion}
Standard language ideology is reflected and reinforced by LLMs powering some of the largest generative AI applications used today, resulting in what we term \textit{standard AI-generated language}. Our faceted taxonomy
illustrates how generative AI systems reproduce standard language ideology---including through normalized “standard” defaults, unequal quality of service, stereotyping and racialization, appropriation and commodification, and suppression or erasure---alongside the broader societal implications. While tensions remain regarding what “desirable” or “appropriate” system behavior looks like, our analysis shows that these issues cannot be resolved through  technical fixes alone; they reflect deeper power relations and institutional priorities. We argue for a shift toward more emancipatory outcomes that foreground accountability, community agency, control, and ownership. To that end, we present three directions for action. First, evaluation frameworks must expand to treat language variety harms as core concerns and meaningful assessments for model performance. Second, researchers, funders, and institutions should support speakers of different language communities in building their own datasets and models, resisting extractive approaches. Third, developers of dominant AI systems must adopt participatory and community-governed approaches that meaningfully redistribute power, rather than reinforcing existing linguistic hierarchies. These approaches go hand-in-hand with rethinking common practices that overrepresent standard varieties in training data through decisions about how data is collected, filtered, and amplified (Section \ref{sec:related-work}). Together, these interventions move toward structural change, recognizing linguistic diversity not as a technical challenge to be managed, but as something to be protected, valued, and sustained---essential for a just AI future. Future work can test, contest, and extend this taxonomy and the recommendations across diverse linguistic, geopolitical, and colonial contexts, and in partnership with the communities most affected. 


\section*{Generative AI Usage Statement}
Generative AI (Gemini) was used for assisting with grammar and fluency of writing and formatting the table.


\bibliography{acmart}


\end{document}